\documentclass[preprint,12pt]{elsarticle}

\usepackage{amssymb}
\usepackage{amsmath}
\usepackage{amsfonts}
\usepackage{algorithmic}
\usepackage{algorithm}
\usepackage{array}
\usepackage[caption=false,font=normalsize,labelfont=sf,textfont=sf]{subfig}
\usepackage{textcomp}
\usepackage{stfloats}
\usepackage{url}
\usepackage{verbatim}
\usepackage{graphicx}
\usepackage{amssymb,amsthm,mathtools}
\usepackage{enumitem}
\usepackage{newfloat}
\usepackage{listings}
\usepackage{multirow}
\usepackage{booktabs}
\usepackage{threeparttable}
\usepackage{capt-of}
\floatstyle{ruled}
\newfloat{listing}{tb}{lst}{}
\floatname{listing}{Listing}
\newtheorem{theorem}{Theorem}
\newtheorem{corollary}[theorem]{Corollary}
\journal{Neurocomputing}
\begin{document}

\begin{frontmatter}



\title{Context on the Imaginary Axis: Input-Conditioned Convolution for Natural Language Inference}

\author{Xin Zhang}
\ead{xin.zhang@ttu.edu}
\author{Hanyao Zeng}
\ead{hazeng@ttu.edu}
\author{Victor S. Sheng\corref{cor1}}
\ead{victor.sheng@ttu.edu}
\cortext[cor1]{Corresponding author}

\affiliation{organization={Department of Computer Science, Texas Tech University},
	city={Lubbock},
	postcode={79409},
	state={TX},
	country={USA}}

\begin{abstract}
We propose Input-Conditioned Complex Convolution (IC$^{3}$), an attention-free method for natural language inference. Building on complex token representations whose imaginary part carries context, IC$^{3}$ generates both a value sequence and a convolution kernel from each premise--hypothesis pair. The kernel determines how the values are combined across the sequence, and the real part of the complex convolution is used to update the token representations. Zero-padded Fourier transforms compute this convolution exactly. Our analysis characterizes the resulting input-dependent operators, establishes limits on their approximation using fixed operators, and separates the effects of changes in the values and the kernel. With all models trained from scratch under a deterministic training protocol over ten random seeds, IC$^{3}$ achieves higher mean accuracy than baseline decoder architectures with comparable parameter counts on ANLI and FOLIO, with consistent gains on SNLI using reduced training data. Controlled operator comparisons and interventions further support the contributions of input conditioning and the imaginary component to predictive performance.
\end{abstract}


\end{frontmatter}

\section{Introduction}

Natural language inference (NLI) requires a model to determine the relation between a premise and a hypothesis, typically among entailment, contradiction, and neutral \citep{bowman2015large}. This task is not only a sentence-level classification problem, but also a structured sentence-pair reasoning problem, in that the model must identify local token-level cues, align related expressions across the two sentences, detect mismatches or contradictions, and compose these local semantics into an overall inference decision. Prior NLI models have therefore emphasized cross-sentence attention, local inference modeling, and inference composition \citep{parikh2016decomposable,chen2017enhanced}. Interaction-space models further show that dense premise--hypothesis interaction tensors contain rich semantic information for solving NLI \citep{gong2018natural}. These observations motivate the joint modeling of token-level semantics and global contextual semantics to determine the relation between a premise and a hypothesis.

Most Transformer-based models represent each token with a real-valued hidden state and learn token--pair interactions implicitly through stacked self-attention layers \citep{vaswani2017attention}. Although self-attention is highly effective, token-level semantic content, contextual information, and premise--hypothesis interaction signals are encoded in the same hidden representation and updated through repeated pairwise token routing. Self-attention provides a general-purpose interaction mechanism, but standard Transformer hidden states do not explicitly separate token-semantic information from complementary contextual information. This raises a central question. \textbf{How can we design an input-conditioned layer that explicitly couples token and context representations for NLI?}

Designing such a layer is challenging because the relation between a premise and a hypothesis can depend on diverse token- and phrase-level semantic cues. The final label may depend on lexical alignment, phrase-level mismatch, negation, quantification, entity substitution, or condition changes. These phenomena motivate integrating token-level information with the context of the full sentence pair during representation updates. Existing interaction-based NLI models address this problem through explicit alignment or interaction spaces \citep{parikh2016decomposable,gong2018natural}, while sequential inference models compose local inference information into an overall decision \citep{chen2017enhanced}. Separately, spectral methods provide efficient alternatives to self-attention, ranging from fixed Fourier mixing and learned global filters to input-dependent frequency filtering \citep{lee2022fnet,rao2021global,huang2023adaptive}. These approaches do not explicitly couple separately represented token and context components through an input-conditioned convolution. We therefore focus on this coupling for NLI.

\begin{figure*}[t]
	\centering
	\includegraphics[width=\linewidth]{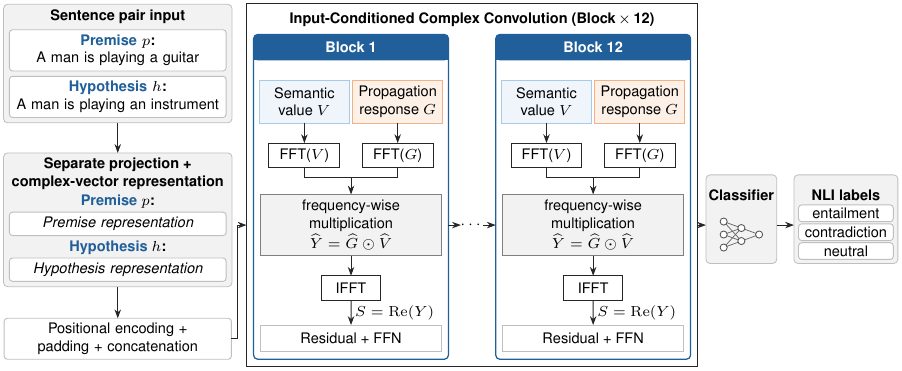}
	\caption{Input-conditioned complex convolution for natural language inference.}
	\label{fig:whole}
\end{figure*}

To address this problem, we propose Input-Conditioned Complex Convolution (\textbf{IC}$^{3}$) for NLI. Figure~\ref{fig:whole} provides an overview of the architecture. Given complex--vector token representations of a premise--hypothesis pair, IC$^{3}$ generates a complex value sequence and a complex convolution kernel from the same input. Their convolution mixes token and context information, and its real part updates the token representations. We adopt complex--vector token representations \citep{zhang2026token} because they place token information in the real part and context in the imaginary part.

This paper makes four contributions. \textbf{First}, we propose \textbf{IC}$^{3}$, an attention-free layer that couples the token and context parts through input-conditioned complex convolution implemented with zero-padded Fourier transforms. \textbf{Second}, we characterize the operators IC$^{3}$ produces, show that they vary in as many directions as their kernels, bound how well fixed operators can approximate them, and separate the output effects of value and kernel changes. \textbf{Third}, we measure how much the learned kernels vary across inputs and test, by projecting, swapping, and editing kernels, whether prediction depends on this variation and on the imaginary part. \textbf{Finally}, we compare \textbf{IC}$^{3}$ with decoder-based baselines of comparable parameter counts, training all models from scratch under a common deterministic protocol over ten random seeds and obtaining higher mean accuracy across the five evaluation settings.

\section{Problem Formulation}

Given a natural language inference instance, let $x=[p;h]$ denote the concatenated premise--hypothesis input, where $p$ is the premise and $h$ is the hypothesis. The goal is to predict the inference relation $y \in \{$entailment, neutral, contradiction$\}$. This decision may depend on token- and phrase-level semantic cues interpreted in the context of the full sentence pair.

Following Zhang and Sheng~\cite{zhang2026token}, we construct Complex--Vector Token Representations separately for the premise and hypothesis before padding and concatenation. Let $L$ and $d$ denote the resulting sequence length and feature dimension. For $0\leq t<L$,
$
X_t=X_t^r+iX_t^i\in\mathbb{C}^d,
$
where $X_t^r,X_t^i\in\mathbb{R}^d$ carry token and context information, respectively. The construction is given in \ref{app:complex_representation}.

Our objective is to learn an input-conditioned convolution that couples these components and updates the real component for NLI.

\section{Method}

Given the complex--vector input representation, \textbf{IC}$^{3}$ uses two branches to generate a complex semantic value sequence and an input-conditioned complex impulse response. For a fixed input, the response is a channel-wise convolution kernel for the value sequence. After zero-padding both sequences to length $2L$, we apply Fourier transforms along the token dimension and multiply the resulting coefficients element-wise. We then apply the inverse Fourier transform, retain the first $L$ positions, and use the real part of the result to update the real component.

\subsection{Semantic and Propagation Response Branches}

Given the complex--vector input representation $X=X^r+iX^i$, where $X^r,X^i\in\mathbb{R}^{L\times d}$, we apply separate layer normalization to the real and imaginary components:
$
\bar{X}^r=\mathrm{LN}_r(X^r),
\;
\bar{X}^i=\mathrm{LN}_i(X^i).
$
Layer normalization acts along the feature dimension at each token position. The normalized components are used to construct a semantic value sequence and an input-conditioned propagation response. All projection matrices belong to $\mathbb{R}^{d\times d}$, all bias vectors belong to $\mathbb{R}^d$, and their parameters are shared across token positions. In the token-wise equations below, individual token representations are written as column vectors.

The semantic value branch produces
$
V=V^r+iV^i\in\mathbb{C}^{L\times d}.
$
For $0\leq t<L$, its components are
$
V_t^r=W_V^r\bar{X}_t^r+b_V^r,
\;
V_t^i=W_V^i\bar{X}_t^i+b_V^i.
$
Thus, the real and imaginary value components are obtained from the token and contextual components, respectively.

The propagation response branch produces
$
G=G^r+iG^i\in\mathbb{C}^{L\times d},
$
with
$
G_t^r=\sigma(W_G^r\bar{X}_t^r+b_G^r),
\;
G_t^i=\tanh(W_G^i\bar{X}_t^i+b_G^i),
\;
0\leq t<L.
$
Here, $\sigma$ denotes the logistic sigmoid function, and both activation functions are applied element-wise. The entries of $G^r$ lie in $(0,1)$, while those of $G^i$ lie in $(-1,1)$.

Although $G_k$ is generated from input position $k$, it is used as the vector of channel-wise coefficients for relative lag $k$ in the subsequent convolution. The same coefficients therefore apply to all source--target pairs with that lag. Thus, $G$ defines an input-conditioned complex impulse-response sequence. 

\subsection{Zero-Padded Spectral Propagation}

Let $\mathcal{F}_N$ denote the length-$N$ discrete Fourier transform (DFT) along the token dimension, applied independently to each feature channel, and let $\mathcal{F}_N^{-1}$ denote its inverse. We use an unnormalized forward DFT and an inverse DFT with normalization factor $1/N$. The semantic value sequence $V$ and the impulse-response sequence $G$ are zero-padded to length $N=2L$, with $(\widetilde{V}_t,\widetilde{G}_t)=(V_t,G_t)$ for $0\leq t<L$ and $(\widetilde{V}_t,\widetilde{G}_t)=(0,0)$ for $L\leq t<N$.

The Fourier representations are
$
\widehat{V}=\mathcal{F}_N(\widetilde{V}),
\;
\widehat{G}=\mathcal{F}_N(\widetilde{G}).
$
Here, $G$ is the complex impulse-response sequence in the token domain, while $\widehat{G}$ is its frequency response. Propagation is computed by
$
\widehat{Y}=\widehat{G}\odot\widehat{V},
$
where $\odot$ denotes element-wise complex multiplication at each frequency and feature channel.

We apply the inverse DFT and retain the first $L$ token positions:
$
\widetilde{Y}=\mathcal{F}_N^{-1}(\widehat{Y}),
\;
Y=P_L\widetilde{Y},
$
where $P_L$ denotes truncation to those positions. Since $N=2L\geq 2L-1$, the zero-padding prevents circular wrap-around. Consequently, $Y$ contains the first $L$ terms of the linear convolution of $G$ and $V$:
$
Y_t=
\sum_{\tau=0}^{t}
G_{t-\tau}\odot V_{\tau},
\;
0\leq t<L.
$
For a fixed input, this defines a channel-wise one-sided convolution of the value sequence. With fast Fourier transforms, it costs $O(dL\log L)$ per layer, compared with $O(dL^2)$ for self-attention.

The real-valued signal used for the token update is
$
S=\operatorname{Re}(Y).
$
Expanding the complex convolution gives
$
S_t
=
\operatorname{Re}(Y_t)
=
\sum_{\tau=0}^{t}
G^r_{t-\tau}\odot V_{\tau}^r
-
\sum_{\tau=0}^{t}
G^i_{t-\tau}\odot V_{\tau}^i,
\;
0\leq t<L.
$
The imaginary response and value components therefore contribute to the token update.

\subsection{Real-Valued Token Update}

For each token position $0\leq t<L$, the real-valued propagation output $S_t$ is projected and normalized:
$
\tilde{S}_t=\mathrm{LN}_o(W_oS_t+b_o),
$
where $W_o\in\mathbb{R}^{d\times d}$ and $b_o\in\mathbb{R}^d$ are shared across token positions. The real component is then updated through a residual connection:
$
R=X^r+\mathrm{Dropout}(\tilde{S}).
$
A position-wise feed-forward network provides a second residual update:
$
X_{\mathrm{out}}^r=R+\mathrm{FFN}(R).
$

The imaginary component is preserved:
$
X_{\mathrm{out}}^i=X^i.
$
The resulting complex hidden state is
$
X_{\mathrm{out}}=X_{\mathrm{out}}^r+iX_{\mathrm{out}}^i.
$
Contextual information contributes to the real-valued update through $S$.

\subsection{Stacked Propagation Layers}

The complex--vector representation is constructed once before the first propagation layer. Each layer receives the preceding complex hidden state and uses its own projections to generate $V$ and $G$. The quantities $\eta_t$, $\alpha_t$, and $\gamma_t$ of \ref{app:complex_representation} are not recomputed between layers. Although the imaginary component is passed unchanged through the stack, each layer applies its own projections to this component when constructing the value and response sequences.

A shared linear classifier maps the real component at the last position of every layer to the three labels. Because the convolution is one-sided, this position receives contributions from every token of the pair. Training minimizes the cross-entropy averaged over all layers, and predictions use the last layer.


\section{Theoretical Analysis}
\label{sec:theory}

This section analyzes the propagation operator induced by the response branch. We first characterize the exact conditional operator class, then quantify how much the operator varies across inputs, and finally separate changes in the propagated values from changes in the propagation operator. Theorem~\ref{thm:family_dimension} and Corollary~\ref{cor:empirical_spectrum} lead to the response-spectrum and projection tests in Section~\ref{sec:mech}, and Theorem~\ref{thm:variation_decomposition} and Corollary~\ref{cor:response_swap} lead to the response-swap test.

\subsection{Conditional Propagation Operator}

Let $\mathcal{X}$ denote the set of admissible layer inputs. For $X\in\mathcal{X}$, let $G(X),V(X)\in\mathbb{C}^{L\times d}$ be the propagation response and semantic value matrices defined in the preceding section. For channel $c\in\{1,\ldots,d\}$, write $g_c(X)=G_{:,c}(X)$ and $v_c(X)=V_{:,c}(X)$.

Let $\mathsf{J}_L\in\mathbb{R}^{L\times L}$ be the truncated shift matrix defined by
\[
(\mathsf{J}_L u)_t=
\begin{cases}
	0, & t=0,\\
	u_{t-1}, & 1\leq t<L.
\end{cases}
\]
For a response vector $g=(g_0,\ldots,g_{L-1})^{\top}\in\mathbb{C}^{L}$, define
$
\Phi_L(g)=\sum_{k=0}^{L-1}g_k\mathsf{J}_L^k.
$
The channel-wise operator induced by the current input is
$
A_c(X)=\Phi_L(g_c(X)).
$
To represent all channels jointly, let $\mathbf v(X)=\operatorname{vec}(V(X))$, $\mathbf y(X)=\operatorname{vec}(Y(X))$, $\boldsymbol{\Phi}_{L,d}(G)=\operatorname{diag}\bigl(\Phi_L(G_{:,1}),\ldots,\Phi_L(G_{:,d})\bigr)$, and $\mathbf A(X)=\boldsymbol{\Phi}_{L,d}(G(X))$. Here, $\operatorname{vec}$ stacks the columns of a matrix in channel order, and $\operatorname{diag}$ forms a block-diagonal matrix from the listed matrix blocks.
The propagation stage can then be written as
$
\mathbf y(X)=\mathbf A(X)\mathbf v(X).
$

\subsection{Shift-Commuting Characterization}

Define the commutant of the truncated shift by
\[
\mathcal{C}_L=
\left\{
B\in\mathbb{C}^{L\times L}:B\mathsf{J}_L=\mathsf{J}_L B
\right\},
\]
and define its channel-wise extension by
\[
\mathcal{C}_{L,d}=
\left\{
\operatorname{diag}(B_1,\ldots,B_d):B_c\in\mathcal{C}_L
\right\}.
\]

\begin{theorem}[Shift-commuting characterization]
	\label{thm:shift_characterization}
	The map $\Phi_L:\mathbb{C}^{L}\to\mathcal{C}_L$ is a complex-linear isomorphism. Equivalently, a matrix $B\in\mathbb{C}^{L\times L}$ commutes with $\mathsf{J}_L$ if and only if there is a unique vector $g\in\mathbb{C}^{L}$ such that
	$
	B=\sum_{k=0}^{L-1}g_k\mathsf{J}_L^k.
	$
	Such a matrix is lower triangular and Toeplitz, and its action on $u\in\mathbb{C}^{L}$ is
	$
	(Bu)_t=\sum_{\tau=0}^{t}g_{t-\tau}u_{\tau},
	\text{ for }0\leq t<L.
	$
	Consequently, $\boldsymbol{\Phi}_{L,d}:\mathbb{C}^{L\times d}\to\mathcal{C}_{L,d}$ is also a complex-linear isomorphism, and
	$
	\dim_{\mathbb{C}}(\mathcal{C}_{L,d})=Ld.
	$
	In particular, every target operator map $\mathbf A_{\mathrm{tar}}:\mathcal{X}\to\mathcal{C}_{L,d}$ admits a unique response map $G_{\mathrm{tar}}:\mathcal{X}\to\mathbb{C}^{L\times d}$ satisfying
	$
	\mathbf A_{\mathrm{tar}}(X)
	=
	\boldsymbol{\Phi}_{L,d}(G_{\mathrm{tar}}(X))
	$
	for every $X\in\mathcal{X}$.
\end{theorem}

Theorem~\ref{thm:shift_characterization} establishes a bijection between unrestricted complex response sequences and channel-wise operators that commute with the truncated shift. The implemented response generator selects a subset of this class through its parameterization and bounded activations. Since $\Phi_L(g)u$ gives the first $L$ entries of the linear convolution of two length-$L$ sequences, zero-padding to any transform length at least $2L-1$ yields an exact Fourier implementation.

\subsection{Dimension of the Conditional Operator Family}
We next quantify how much the selected operator can vary with the input. We measure affine dimension over $\mathbb{R}$ by jointly representing the real and imaginary components of each complex response as a real vector. This convention matches the real-valued coefficients used in the affine approximations below.

Fix a reference input $X_{\circ}\in\mathcal{X}$. Let $\mathcal{G}=\{G(X):X\in\mathcal{X}\}$ denote the family of input-conditioned propagation responses, and let $\mathcal{A}=\{\mathbf A(X):X\in\mathcal{X}\}$ denote the corresponding family of conditional propagation operators. Define
$
n_{\mathcal{G}}=\dim_{\mathbb{R}}\operatorname{span}_{\mathbb{R}}\{G(X)-G(X_{\circ}):X\in\mathcal{X}\},
$
which is the real affine dimension of the response family $\mathcal{G}$. Equivalently, $n_{\mathcal{G}}$ is the minimum number of fixed response directions needed to express all differences $G(X)-G(X_{\circ})$ as linear combinations with real coefficients.
For the quantitative statement, define
$
\mathsf{Q}_L=\operatorname{diag}(\sqrt{L},\sqrt{L-1},\ldots,1)
$
and the weighted realification map
$
\omega(H)=
\begin{bmatrix}
	\operatorname{vec}(\operatorname{Re}(\mathsf{Q}_L H))\\
	\operatorname{vec}(\operatorname{Im}(\mathsf{Q}_L H))
\end{bmatrix}
\in\mathbb{R}^{2Ld}.
$
Let $\boldsymbol{\xi}(X)=\omega(G(X))$. When $X$ is random, assume $\mathbb{E}\|\boldsymbol{\xi}(X)\|_2^2<\infty$, let $\boldsymbol{\mu}=\mathbb{E}[\boldsymbol{\xi}(X)]$, and define
$
\boldsymbol{\Sigma}=
\mathbb{E}
\left[
(\boldsymbol{\xi}(X)-\boldsymbol{\mu})
(\boldsymbol{\xi}(X)-\boldsymbol{\mu})^{\top}
\right].
$
Let $q=2Ld$ and let $\lambda_1\geq\cdots\geq\lambda_q\geq0$ be the eigenvalues of $\boldsymbol{\Sigma}$.
For an operator $C\in\mathcal{C}_{L,d}$ and a nonempty real affine subspace $\mathcal{U}\subseteq\mathcal{C}_{L,d}$, define
$
\operatorname{dist}_{\mathrm F}(C,\mathcal{U})
=
\inf_{D\in\mathcal{U}}\|C-D\|_{\mathrm F},
$
where $D$ ranges over the operators in $\mathcal{U}$ and $\|\cdot\|_{\mathrm F}$ denotes the Frobenius norm.

\begin{theorem}[Dimension preservation and finite-dictionary approximation]
	\label{thm:family_dimension}
	The response family and the induced operator family have the same real affine dimension:
	$
	\operatorname{affdim}_{\mathbb{R}}(\mathcal{A})
	=
	\operatorname{affdim}_{\mathbb{R}}(\mathcal{G})
	=
	n_{\mathcal{G}}.
	$
	Moreover, the weighted coordinate map is isometric with respect to the operator Frobenius norm:
	$
	\left\|
	\boldsymbol{\Phi}_{L,d}(H_1)-\boldsymbol{\Phi}_{L,d}(H_2)
	\right\|_{\mathrm F}^2
	=
	\left\|
	\omega(H_1)-\omega(H_2)
	\right\|_2^2
	$
	for all $H_1,H_2\in\mathbb{C}^{L\times d}$.
	
	For every integer $K$ satisfying $0\leq K<q$, the best $K$-dimensional affine approximation of the random operator family satisfies
	\[	\inf_{\substack{
			\mathcal{U}\subseteq\mathcal{C}_{L,d},\ \mathcal{U}\text{ affine}\\
			\operatorname{affdim}_{\mathbb{R}}(\mathcal{U})\leq K
	}}
	\mathbb{E}
	\left[
	\operatorname{dist}_{\mathrm F}(\mathbf A(X),\mathcal{U})^2
	\right]
	=
	\sum_{j=K+1}^{q}\lambda_j.
	\]
	Consequently, any dictionary-conditioned operator of the form
	$
	\widetilde{\mathbf A}(X)=B_0+\sum_{j=1}^{K}\beta_j(X)B_j,
	$
	where $B_0,B_1,\ldots,B_K\in\mathcal{C}_{L,d}$ are fixed and $\beta_j(X)\in\mathbb{R}$, obeys
	$
	\mathbb{E}
	\left[
	\|\mathbf A(X)-\widetilde{\mathbf A}(X)\|_{\mathrm F}^2
	\right]
	\geq
	\sum_{j=K+1}^{q}\lambda_j.
	$
	
	Thus, an approximation using $K$ fixed operator directions cannot represent $\mathbf A(X)$ exactly for all $X\in\mathcal{X}$ when $n_{\mathcal{G}}>K$. Similarly, an affine mixture of $K\geq1$ fixed expert operators has the form
	$
	\widetilde{\mathbf A}_{\mathrm{mix}}(X)=\sum_{j=1}^{K}\pi_j(X)E_j,
	$
	where $\pi_j(X)\in\mathbb{R}$ and $\sum_{j=1}^{K}\pi_j(X)=1$. Its range has real affine dimension at most $K-1$, so it cannot represent the entire conditional operator family when $n_{\mathcal{G}}>K-1$.
\end{theorem}

\begin{corollary}[Empirical response spectrum]
	\label{cor:empirical_spectrum}
	Let $X_1,\ldots,X_{n_{\mathrm{s}}}$ be observed inputs, let
	$
	\overline{\boldsymbol{\xi}}=
	\frac{1}{n_{\mathrm{s}}}
	\sum_{m=1}^{n_{\mathrm{s}}}\boldsymbol{\xi}(X_m),
	$
	and form the centered matrix $\mathbf M_{\xi}\in\mathbb{R}^{n_{\mathrm{s}}\times q}$ whose $m$-th row is $(\boldsymbol{\xi}(X_m)-\overline{\boldsymbol{\xi}})^{\top}$. If $s_1\geq\cdots\geq s_{\min\{n_{\mathrm{s}},q\}}\geq0$ are the singular values of $\mathbf M_{\xi}$, then, for every integer $0\leq K<q$, the minimum average squared Frobenius distance from $\mathbf A(X_1),\ldots,\mathbf A(X_{n_{\mathrm{s}}})$ to a real affine operator subspace of dimension at most $K$ is
	$
	\frac{1}{n_{\mathrm{s}}}
	\sum_{j=K+1}^{\min\{n_{\mathrm{s}},q\}}s_j^2,
	$
	where the sum is interpreted as zero when $K\geq\min\{n_{\mathrm{s}},q\}$.
	In particular, $\operatorname{rank}(\mathbf M_{\xi})$ is the empirical affine dimension of the sampled conditional operator family.
\end{corollary}

\subsection{Value and Operator Variation}

For a response matrix $G\in\mathbb{C}^{L\times d}$, define its maximum channel-wise $\ell_1$ norm by
$
\|G\|_1=
\max_{1\leq c\leq d}
\sum_{k=0}^{L-1}|G_{k,c}|.
$

\begin{theorem}[Value--operator decomposition and stability]
	\label{thm:variation_decomposition}
	For any $X,X^{\prime}\in\mathcal{X}$,
	$
	\mathbf y(X^{\prime})-\mathbf y(X)
	={}
	\mathbf A(X^{\prime})
	\bigl(\mathbf v(X^{\prime})-\mathbf v(X)\bigr)
	+
	\bigl(\mathbf A(X^{\prime})-\mathbf A(X)\bigr)
	\mathbf v(X).
	$
	The first term is the value-variation term under the operator selected by $X^{\prime}$, whereas the second term is the operator-variation term acting on the value sequence from $X$. The decomposition implies
	$
	\|\mathbf y(X^{\prime})-\mathbf y(X)\|_2
	\leq{}
	\|G(X^{\prime})\|_1
	\|\mathbf v(X^{\prime})-\mathbf v(X)\|_2
	+
	\|G(X^{\prime})-G(X)\|_1
	\|\mathbf v(X)\|_2.
	$
	If the response is input independent, so that $G(X)=G_0$ for all $X$, then $\mathbf A(X)=\mathbf A_0$ and the operator-variation term vanishes.
\end{theorem}

\begin{corollary}[Response-swap intervention]
	\label{cor:response_swap}
	Define the swapped-response output by
	$
	\mathbf y(X\mid X^{\prime})
	=
	\mathbf A(X^{\prime})\mathbf v(X).
	$
	Then
	$
	\mathbf y(X\mid X^{\prime})-\mathbf y(X\mid X)
	=
	\bigl(\mathbf A(X^{\prime})-\mathbf A(X)\bigr)\mathbf v(X).
	$
	Thus, swapping the response while holding the value sequence fixed isolates the operator-variation term in Theorem~\ref{thm:variation_decomposition} exactly at the propagation output.
\end{corollary}

\subsection{Relation to Existing Sequence Operators}

The preceding results clarify the relation between IC$^{3}$ and several existing operator constructions. A static convolution or fixed spectral mixer uses an input-independent propagation operator. Its operator family therefore has affine dimension zero, and the operator-variation term in Theorem~\ref{thm:variation_decomposition} vanishes. Conditional combinations of fixed kernels, as in dynamic convolution \citep{chen2020dynamic}, restrict operator variation to the affine space determined by those kernels. Theorem~\ref{thm:family_dimension} quantifies the approximation error when the IC$^{3}$ operator family is represented using finitely many fixed directions in $\mathcal{C}_{L,d}$. Direct filter generation, as in Dynamic Filter Networks \citep{jia2016dynamic}, also produces input-dependent operators.

IC$^{3}$ combines input conditioning with a channel-wise, one-sided convolution structure. For each fixed input, the propagation coefficients within each channel are shared across token pairs with the same lag, yielding a lower-triangular Toeplitz operator that commutes with $\mathsf{J}_L$. Position-dependent dynamic convolution \citep{wu2019pay} and dense attention \citep{vaswani2017attention} need not satisfy this constraint. Consequently, the IC$^{3}$ propagation operator is specified by $Ld$ complex lag coefficients, whereas an unrestricted channel-wise complex source--target operator requires $L^2d$ coefficients.


\begin{table*}[t]
	\centering
	\caption{Performance comparison of \textbf{IC}$^{3}$ and same-scale baseline models over 10 seeds.}
	\label{tab:inference_results}
	\scriptsize
	\setlength{\tabcolsep}{2.0pt}
	\renewcommand{\arraystretch}{0.95}
	\begin{threeparttable}
			\begin{tabular}{@{}lcccccc@{}}
				\toprule
				\textbf{Model} &
				\textbf{Metric} &
				\textbf{ANLI-R1} &
				\textbf{ANLI-R2} &
				\textbf{ANLI-R3} &
				\textbf{FOLIO} &
				\textbf{SNLI (5\%)}  \\
				\midrule
				
				\textbf{IC}$^{3}$ &
				Acc \% &
				\textbf{39.01} &
				\textbf{46.08} &
				\textbf{44.87} &
				\textbf{50.34} &
				\textbf{63.86} \\
				&
				$\Delta$ CI &
				\textemdash &
				\textemdash &
				\textemdash &
				\textemdash &
				\textemdash \\
				
				\midrule
				Llama 3.2 &
				Acc \% &
				36.96 &
				43.61 &
				42.48 &
				45.22 &
				59.15 \\
				&
				$\Delta$ CI &
				2.05 [0.70,3.40] &
				2.47 [1.63,3.31] &
				2.38 [1.04,3.72] &
				5.12 [2.16,8.09] &
				4.71 [4.30,5.11] \\
				
				\midrule
				Phi 1.5 &
				Acc \% &
				37.66 &
				43.57 &
				41.74 &
				44.19 &
				61.80 \\
				&
				$\Delta$ CI &
				1.35 [0.12,2.58] &
				2.51 [1.50,3.52] &
				3.12 [2.36,3.88] &
				6.16 [3.27,9.04] &
				2.06 [1.76,2.37] \\
				
				\midrule
				Pythia &
				Acc \% &
				36.09 &
				39.07 &
				37.22 &
				37.59 &
				59.83 \\
				&
				$\Delta$ CI &
				2.92 [2.22,3.62] &
				7.01 [5.06,8.96] &
				7.65 [4.77,10.53] &
				12.76 [9.81,15.70] &
				4.03 [3.56,4.50] \\
				
				\midrule
				Gemma 3 &
				Acc \% &
				36.77 &
				41.51 &
				41.73 &
				43.69 &
				60.99 \\
				&
				$\Delta$ CI &
				2.24 [1.46,3.02] &
				4.57 [3.84,5.30] &
				3.14 [2.09,4.19] &
				6.65 [4.22,9.08] &
				2.87 [2.19,3.55] \\
				\bottomrule
			\end{tabular}%
		
		\begin{tablenotes}
			\scriptsize
			\item Acc denotes accuracy. $\Delta$ denotes the mean paired accuracy difference between \textbf{IC}$^{3}$ and each baseline over ten matched random seeds. Brackets report two-sided 95\% paired Student's $t$ confidence intervals. All values are percentages.
		\end{tablenotes}
		
	\end{threeparttable}
\end{table*}
\section{Results}
Our experiments evaluate predictive performance and the role of input-conditioned propagation through benchmark comparisons, architectural ablations, response-spectrum analysis, response projection, response swapping, imaginary-response interventions, and input-order controls. We use FP32 training with deterministic PyTorch and CUDA settings, with TF32 and cuDNN benchmarking disabled. For each architecture and dataset, learning rate and dropout are selected by mean validation performance over three seeds, followed by evaluation on seven additional seeds. We report ten-seed test means and paired $95\%$ Student's $t$ CIs for model comparisons, and across-seed means and SDs ($\pm$) for intervention analyses unless otherwise noted. All models are trained on a GB200/GH200 platform with 128GB memory and 8 CPU cores.

We train \textbf{IC}$^{3}$ and same-scale decoder-based baselines from scratch on the same data. Our model uses 12 layers with a 1536-dimensional embedding size, resulting in 0.57B parameters. It uses the Llama 3.2 tokenizer and token embeddings learned from scratch, and it is trained with AdamW, batch size 16, linear warm-up followed by cosine decay, gradient clipping at 1.2, and early stopping on validation accuracy within 30 epochs. To ensure a comparable parameter budget, we reduce the number of layers in the 1B baseline architectures and compare all models at approximately the same parameter scale. The baselines are adapted from Llama 3.2-1B \citep{grattafiori2024llama}, Phi-1.5 \citep{li2023textbooks}, Pythia 1B \citep{biderman2023pythia}, and Gemma 3-1B \citep{gemma2025gemma3}. They use their native tokenizers, the Hugging Face Trainer, and a linear classifier on the last token of the final layer. All compared models are trained using matched random seeds, deterministic training, and the same validation-based hyperparameter selection protocol.

We evaluate on ANLI-R1, ANLI-R2, and ANLI-R3 \citep{nie2020adversarial} for adversarial NLI, FOLIO \citep{han2024folio} for logically structured natural-language reasoning, and SNLI \citep{bowman2015large} with reduced training data. For SNLI, we sample $5\%$ of the training set and $10\%$ of the validation set, stratified by label, and evaluate on the full test set. 

\paragraph{Decoder comparison}
Table~\ref{tab:inference_results} shows that \textbf{IC$^{3}$ consistently outperforms all same-scale decoder-based baselines across all five evaluation settings}. The positive confidence intervals of $\Delta$ indicate that the gains are stable under the ten-seed evaluation protocol. The largest advantage over every baseline appears on FOLIO. The SNLI result further suggests that the trend extends to a larger NLI dataset under a reduced-data training setting.

\paragraph{Encoder comparison}
Table~\ref{tab:encoder_r1} compares IC$^{3}$ with a randomly initialized ModernBERT-large encoder \citep{warner-etal-2025-smarter} on ANLI-R1. IC$^{3}$ achieves a mean accuracy of $39.01\%$, compared with $35.25\%$ for ModernBERT-large, and the paired $95\%$ confidence interval excludes zero.

\begin{table}[t]
	\centering
	\caption{ANLI-R1 comparison with encoder architectures over ten matched seeds.}
	\label{tab:encoder_r1}
	\small
	\setlength{\tabcolsep}{5pt}
	\renewcommand{\arraystretch}{1.05}
	\begin{tabular}{@{}lccc@{}}
		\toprule
		\textbf{Model} &
		\textbf{Params} &
		\textbf{Acc. (\%)} &
		\textbf{Diff [95\% CI]} \\
		\midrule
		\textbf{IC}$^{3}$ &
		0.57B &
		\textbf{39.01} &
		\textemdash \\
		ModernBERT-large &
		0.40B &
		35.25 &
		3.76 [1.74, 5.78] \\
		\bottomrule
	\end{tabular}
\end{table}

\paragraph{Controlled operator comparison}
Table~\ref{tab:spectral_control} compares IC$^{3}$ with FNet-style \citep{lee2022fnet}, Orchid-based \citep{karami2024orchid}, and AFF \citep{huang2023adaptive} mixing under the same ten-seed protocol. These variants share the IC$^{3}$ embedding and training setup, including the layer-averaged loss. IC$^{3}$ achieves higher mean accuracy than all three alternatives on ANLI-R1, ANLI-R2, and ANLI-R3. All paired $95\%$ confidence intervals exclude zero except for Orchid-based mixing on ANLI-R1.

\begin{table*}[t]
	\centering
	\caption{Operator comparison over ten matched seeds. Each baseline entry reports accuracy followed by the paired $\Delta$ and its $95\%$ confidence interval relative to IC$^{3}$.}
	\label{tab:spectral_control}
	\scriptsize
	\setlength{\tabcolsep}{4pt}
	\renewcommand{\arraystretch}{0.95}
	\begin{tabular}{@{}lccc@{}}
		\toprule
		\textbf{Propagation operator} &
		\textbf{ANLI-R1} &
		\textbf{ANLI-R2} &
		\textbf{ANLI-R3} \\
		\midrule
		IC$^{3}$ propagation &
		\textbf{39.01} &
		\textbf{46.08} &
		\textbf{44.87} \\
		FNet-style mixing &
		37.95; $1.06\,[0.04,2.08]$ &
		42.83; $3.25\,[2.45,4.05]$ &
		43.53; $1.34\,[0.72,1.95]$ \\
		Orchid-based mixing &
		38.31; $0.70\,[-0.55,1.95]$ &
		42.98; $3.10\,[1.82,4.38]$ &
		41.53; $3.33\,[2.27,4.40]$ \\
		AFF &
		36.51; $2.50\,[1.58,3.42]$ &
		43.00; $3.08\,[2.42,3.74]$ &
		41.13; $3.73\,[2.66,4.80]$ \\
		\bottomrule
	\end{tabular}%
	
\end{table*}
\paragraph{Ablations}
Table~\ref{tab:ablation} shows that replacing the proposed propagation mechanism, removing the complex token--context representation, or restricting the response substantially degrades performance. Each variant changes only the listed component and keeps all other components and training settings unchanged. In particular, the static-response and imaginary-response ablations support the importance of input conditioning and of the imaginary part. These results support the combined IC$^{3}$ design.

\begin{table*}[t]
	\centering
	
	\begin{minipage}[t]{\linewidth}
		\centering
		\captionof{table}{Component ablations on ANLI-R2}
		\label{tab:ablation}
		\footnotesize
		\setlength{\tabcolsep}{2.5pt}
		\begin{tabular}{@{}lcc@{}}
			\toprule
			Variant & Acc \% & $\Delta$ CI \\
			\midrule
			Full \textbf{IC}$^{3}$ 
			& 46.08 & \textemdash \\
			w/ token--context attention 
			& 42.82 & 3.26 [2.27, 4.25] \\
			w/ real-valued input 
			& 39.29 & 6.79 [4.65, 8.93] \\
			w/ static response 
			& 40.99 & 5.09 [4.29, 5.89] \\
			w/o imaginary response 
			& 39.94 & 6.14 [4.88, 7.40] \\
			\bottomrule
		\end{tabular}
		\end{minipage}
	\par\bigskip
	\begin{minipage}[t]{\linewidth}
		\centering
		\captionof{table}{Layer-0 interventions on ANLI-R2 over seven seeds}
		\label{tab:operator_interventions}
		\footnotesize
		\setlength{\tabcolsep}{3pt}
		\begin{tabular}{@{}lc@{}}
			\toprule
			Intervention & Acc. drop (pp)\\
			\midrule
			Random response swap
			& $10.86 \pm 1.96$ \\
			Same-label response swap
			& $8.21 \pm 1.77$ \\
			Remove imaginary response
			& $12.27 \pm 1.98$ \\
			Reverse imaginary response
			& $20.59 \pm 1.34$ \\
			\bottomrule
		\end{tabular}
		
	\end{minipage}
	
\end{table*}
\subsection{Mechanistic Validation of the Conditional Operator Analysis}\label{sec:mech}
The preceding ablations establish that input conditioning and the imaginary response are useful at the architectural level. We next examine whether the trained model uses the operator variation characterized in Section~\ref{sec:theory}. We ask whether the learned operators vary in many directions, whether this variation matters for prediction, and whether the imaginary response contributes to the token update.

\begin{figure*}[!htbp]
	\centering
	\includegraphics[width=0.85\linewidth]{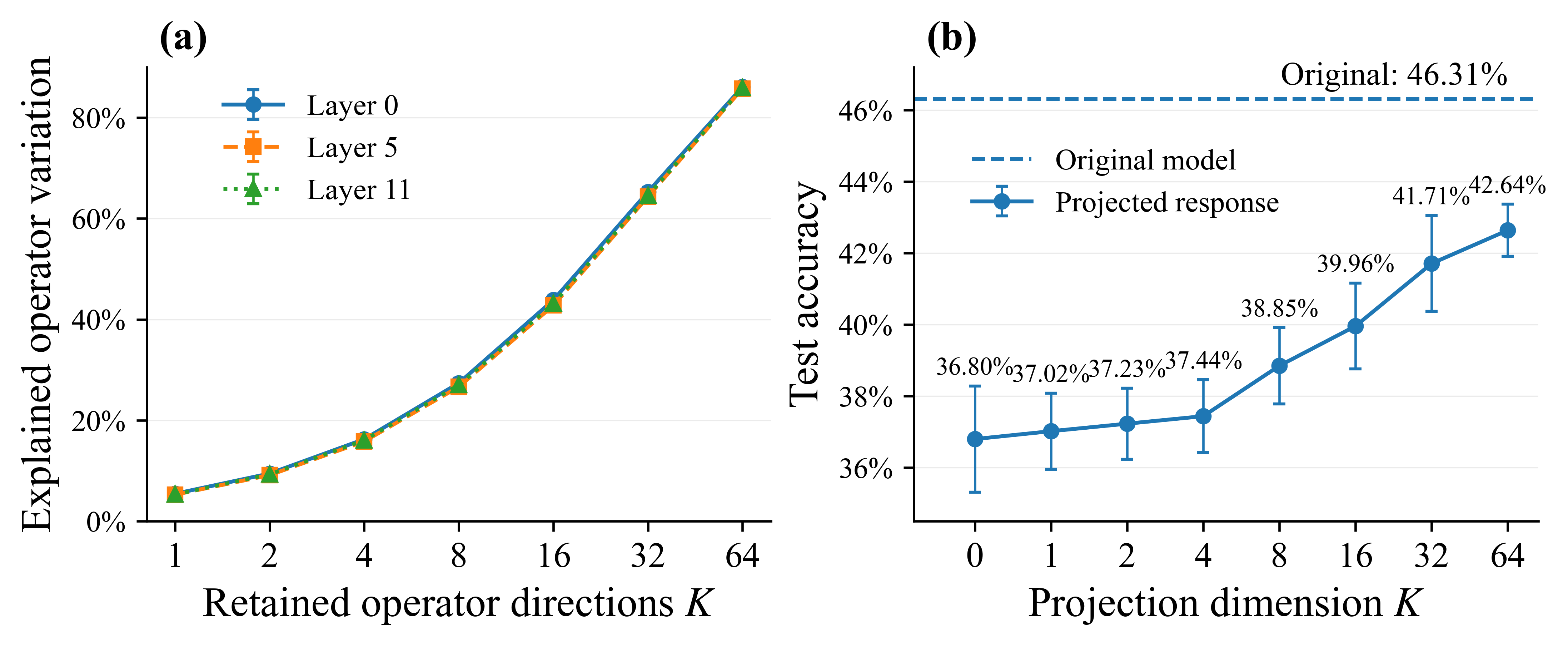}
	\caption{Geometry and task relevance of the learned conditional operator family. Panel (b) uses Layer 0. Original denotes the unmodified model evaluated with the same seven seeds and fixed input length as the intervened models.}
	\label{fig:operator_geometry}
\end{figure*}
\paragraph{Response variation across principal components}
Figure~\ref{fig:operator_geometry}(a) reports the empirical response spectrum characterized by Corollary~\ref{cor:empirical_spectrum}, computed from up to 512 training inputs whose longer sentence has the same number of tokens, so that all responses have the same size. Panel (b) uses test inputs of the same length. At Layer 0, explaining $90\%$ of the observed variation requires $81.3 \pm 2.5$ principal components across seeds, with similar spectra at Layers 5 and 11. The observed response variation is therefore distributed over many principal components at the examined layers.

\paragraph{Predictive performance under response projection}
Figure~\ref{fig:operator_geometry}(b) replaces each Layer-0 response with its projection onto the mean and the top $K$ principal directions. Accuracy rises from $(36.80 \pm 1.48)\%$ at $K=0$ to $(42.64 \pm 0.73)\%$ at $K=64$, versus $(46.31 \pm 0.75)\%$ for the unmodified model on the same test inputs. Replacing the responses of any single layer from 1 to 11 by their mean changes the mean accuracy by at most $0.06$ percentage points. Under these single-layer interventions, accuracy is most sensitive to input-dependent responses at Layer 0. Layer 0 is the first layer in which premise and hypothesis tokens interact, because the complex--vector representations are constructed separately for the two sentences. An input-independent response at this layer therefore affects the input to every later layer. 

\paragraph{Response swapping tests input--operator matching}
Corollary~\ref{cor:response_swap} shows that response swapping isolates the operator-variation term at the propagation output while holding the value sequence fixed at the intervened layer. As summarized in Table~\ref{tab:operator_interventions}, replacing the Layer-0 response with that generated by another input reduces accuracy by $10.86 \pm 1.96$ percentage points, and a substantial degradation remains when the donor input has the same label. Response swapping in later layers has little effect on accuracy. The same-label result suggests that predictive performance depends on matching the Layer-0 response to the input, even when the donor has the same class label.

\paragraph{The imaginary response contributes to the real-valued update}
The complex decomposition shows that the imaginary response and value components contribute jointly to the real-valued propagation output. Removing the Layer-0 imaginary response reduces accuracy by $12.27 \pm 1.98$ percentage points, while reversing its sign produces an even larger degradation, as shown in Table~\ref{tab:operator_interventions}. These interventions show that the trained model uses the imaginary contribution for prediction, and the ablation in Table~\ref{tab:ablation} shows that a model trained without the imaginary response is $6.14$ percentage points less accurate. The spectrum, projection, response-swap, and imaginary-response results provide empirical support for the conditional operator analysis in Section~\ref{sec:theory}.

\paragraph{Effect of input order}
With a one-sided convolution, tokens of the first sentence cannot see the second sentence. Under the ten-seed evaluation protocol, the premise--hypothesis and hypothesis--premise input orders achieve mean accuracies of $46.08\%$ and $45.82\%$ on ANLI-R2, respectively. The mean paired difference is $0.26$ percentage points, with a $95\%$ confidence interval of $[-0.79,1.31]$. The difference is not statistically significant.

\section{Related Work}

\paragraph{Natural language inference and structured sentence-pair reasoning}
NLI benchmarks such as SNLI, MultiNLI, and ANLI evaluate entailment, contradiction, and neutral relations under increasingly challenging settings \citep{bowman2015large,williams2018broad,nie2020adversarial}, while LogicNLI and FOLIO emphasize logically structured reasoning \citep{tian2021diagnosing,han2024folio}. Neural approaches include token alignment in Decomposable Attention \citep{parikh2016decomposable}, local inference composition in ESIM \citep{chen2017enhanced}, dense premise--hypothesis interaction modeling \citep{gong2018natural}, and pretrained cross-encoders such as BERT and DeBERTa \citep{devlin2019bert,he2021deberta}.

\paragraph{Attention-free and spectral sequence operators}
Structured alternatives to self-attention include S4 \citep{gu2022efficiently}, Mamba \citep{gu2023mamba}, RetNet \citep{sun2023retentive}, Gated Linear Attention \citep{3692070.3694403}, Hyena \citep{poli2023hyena}, and lightweight or dynamic convolutions \citep{wu2019pay}. Spectral token mixing includes FNet \citep{lee2022fnet}, GFNet \citep{rao2021global}, AFNO \citep{guibas2022efficient}, Caracal \citep{gan2026caracal}, and Adaptive Frequency Filtering \citep{huang2023adaptive}.

\paragraph{Complex-valued representations}
Complex-valued neural models exploit magnitude and phase structure \citep{lee2022complex}. In language modeling and matching, CNM uses complex semantic representations \citep{li2019cnm}, EulerFormer applies polar-form complex representations \citep{tian2024eulerformer}, and Complex--Vector Token Representations separate token and context information \citep{zhang2026token}.

\paragraph{Input-conditioned operators}
Dynamic Filter Networks generate filters from the current input \citep{jia2016dynamic}, HyperNetworks generate parameters for another network \citep{ha2016hypernetworks}, and dynamic convolution combines kernels using input-dependent weights \citep{chen2020dynamic}. These approaches establish input-conditioned parameterization as a general mechanism for adaptive computation.

IC$^{3}$ differs from fixed spectral mixers such as FNet and GFNet by generating its propagation response from the current premise--hypothesis input. Unlike general dynamic filtering, it restricts the conditional operator to channel-wise one-sided convolution with coefficients shared across equal relative lags. Relative to Complex--Vector Token Representations with Token--Context Attention, it retains the token--context decomposition but replaces attention-based routing with zero-padded spectral convolution.

\section{Conclusion}
We introduced \textbf{IC}$^{3}$, an attention-free NLI architecture with input-conditioned complex convolution. Our analysis characterizes its operators and separates value changes from kernel changes. Experiments show consistent gains over same-scale baselines and confirm that the learned operators are multidimensional, input-dependent, and reliant on the imaginary part.

\subsection*{Reproducibility statement}
The supplementary material provides the source code and complete experimental configurations used to reproduce the experiments and numerical results reported in this paper.

\appendix
\section{Complex--Vector Token Representation}
\label{app:complex_representation}

Following Zhang and Sheng~\cite{zhang2026token}, we project the premise and hypothesis token embeddings with separate linear layers and apply the complex--vector construction to each sequence. The derivation below applies to either sequence. For the $t$-th token, let $a_t \in \mathbb{R}^d$ denote its real-valued token representation. The global semantic vector is computed dimension-wise as $\rho=\sqrt{\sum_{\tau}a_{\tau}\odot a_{\tau}}$, and the contextual representation of token $t$ is constructed from the remaining tokens as $\eta_t=\sqrt{\sum_{\tau\neq t}a_{\tau}\odot a_{\tau}}$, where $\odot$ denotes element-wise multiplication. The token--global alignment phase is defined dimension-wise by
$
\alpha_t=\operatorname{atan2}\left(\sqrt{1-\left(\frac{a_t}{\rho}\right)^2},\frac{a_t}{\rho}\right).
$
This phase characterizes the angular relation between the token representation and the global semantic vector. The polar-form complex representation uses the adapted magnitude $\rho+a_t$ and the phase $\alpha_t$:
$
Z_t=(\rho+a_t)\odot e^{i\alpha_t}.
$
After converting $Z_t$ to Cartesian form, the real and imaginary components can be written as
$
X_t^r=\gamma_t\odot a_t, X_t^i=\gamma_t\odot \eta_t,
$
where $\gamma_t=\cos(\alpha_t)+1$ is the directional gate. Thus, the complex-vector token representation used in this paper is
$
X_t=X_t^r+iX_t^i.
$
Before padding and concatenation, a sinusoidal positional encoding, with positions counted within each sentence, is added to the real component.

\section{Proofs of Theoretical Results}
\label{app:proofs}

This appendix provides the proofs of the theoretical results stated in Section~\ref{sec:theory}, following the same order as in the main text.

\subsection{Proof of Theorem~\ref{thm:shift_characterization}}
\label{app:proof_shift_characterization}

\begin{proof}
	Every polynomial in $\mathsf{J}_L$ commutes with $\mathsf{J}_L$, so $\Phi_L(g)\in\mathcal{C}_L$ for every $g\in\mathbb{C}^{L}$. Conversely, let $B\in\mathcal{C}_L$ and let $e_0,\ldots,e_{L-1}$ be the standard basis of $\mathbb{C}^{L}$. Since $e_t=\mathsf{J}_L^t e_0$, commutation gives
	$
	Be_t=B\mathsf{J}_L^t e_0=\mathsf{J}_L^t Be_0.
	$
	Write $Be_0=\sum_{k=0}^{L-1}g_k e_k$. Then
	$
	Be_t=\sum_{k=0}^{L-1-t}g_k e_{k+t},
	$
	which shows that $B=\sum_{k=0}^{L-1}g_k\mathsf{J}_L^k$. The coefficients are unique because they form the first column of $B$. Since $\Phi_L$ is linear by definition, it is a complex-linear isomorphism. The convolution formula follows by direct multiplication. Applying the same argument independently to each channel proves the block-diagonal statement and the dimension formula. The final claim follows from the bijectivity of $\boldsymbol{\Phi}_{L,d}$.
\end{proof}

\subsection{Proof of Theorem~\ref{thm:family_dimension}}
\label{app:proof_family_dimension}

\begin{proof}
	The map $\boldsymbol{\Phi}_{L,d}$ is injective and real-linear, so it preserves affine independence and therefore preserves affine dimension. For a single channel, the coefficient at lag $k$ appears on a subdiagonal containing $L-k$ entries. Hence
	$
	\|\Phi_L(g)\|_{\mathrm F}^2
	=
	\sum_{k=0}^{L-1}(L-k)|g_k|^2.
	$
	Summing over channels gives the stated isometry.
	
	The isometry reduces affine approximation of $\mathbf A(X)$ under the Frobenius norm to affine approximation of $\boldsymbol{\xi}(X)$ under the Euclidean norm. The optimal affine subspace contains the mean $\boldsymbol{\mu}$, and its optimal $K$ directions are the eigenvectors associated with the largest $K$ eigenvalues of $\boldsymbol{\Sigma}$. The residual expected squared distance is therefore
	$
	\sum_{j=K+1}^{q}\lambda_j.
	$
	
	The range of $\widetilde{\mathbf A}$ is contained in the affine space
	$
	B_0+\operatorname{span}_{\mathbb{R}}\{B_1,\ldots,B_K\},
	$
	whose affine dimension is at most $K$. Applying the optimal affine-approximation identity proves the lower bound. The exact-representation statement follows from affine-dimension preservation. For an affine mixture of $K\geq1$ fixed expert operators with real coefficients summing to one, the range lies in their real affine hull, whose dimension is at most $K-1$. This proves the corresponding exact-representation statement for affine mixtures.
\end{proof}

\subsection{Proof of Corollary~\ref{cor:empirical_spectrum}}
\label{app:proof_empirical_spectrum}

\begin{proof}
	Apply Theorem~\ref{thm:family_dimension} to the empirical distribution supported uniformly on $X_1,\ldots,X_{n_{\mathrm{s}}}$ and use the singular-value characterization of the best rank-$K$ approximation.
\end{proof}

\subsection{Proof of Theorem~\ref{thm:variation_decomposition}}
\label{app:proof_variation_decomposition}

\begin{proof}
	Add and subtract $\mathbf A(X^{\prime})\mathbf v(X)$ to obtain the exact decomposition.
	
	For each channel, extend $g$ and $u$ by zero outside $\{0,\ldots,L-1\}$. Young's convolution inequality and the fact that truncation does not increase the $\ell_2$ norm give
	$
	\|\Phi_L(g)u\|_2
	\leq
	\|g\|_1\|u\|_2.
	$
	Since $\boldsymbol{\Phi}_{L,d}(G)$ is block diagonal, it follows that
	$
	\|\boldsymbol{\Phi}_{L,d}(G)\|_{2\to2}
	\leq
	\max_{1\leq c\leq d}\|G_{:,c}\|_1
	=
	\|G\|_1.
	$
	By linearity,
	$
	\mathbf A(X^{\prime})-\mathbf A(X)
	=
	\boldsymbol{\Phi}_{L,d}\bigl(G(X^{\prime})-G(X)\bigr).
	$
	Applying the operator-norm bound to $G(X^{\prime})$ and $G(X^{\prime})-G(X)$, and then using the triangle inequality in the decomposition, proves the stated stability bound.
	
	If $G(X)=G_0$ for all $X$, then $\mathbf A(X)=\boldsymbol{\Phi}_{L,d}(G_0)$ is constant. Hence $\mathbf A(X^{\prime})-\mathbf A(X)=0$, and the operator-variation term vanishes.
\end{proof}

\subsection{Proof of Corollary~\ref{cor:response_swap}}
\label{app:proof_response_swap}

\begin{proof}
	By definition,
	$
	\mathbf y(X\mid X)=\mathbf A(X)\mathbf v(X)
	$
	and
	$
	\mathbf y(X\mid X^{\prime})=\mathbf A(X^{\prime})\mathbf v(X).
	$
	Subtracting gives
	$
	\mathbf y(X\mid X^{\prime})-\mathbf y(X\mid X)
	=
	\bigl(\mathbf A(X^{\prime})-\mathbf A(X)\bigr)\mathbf v(X).
	$
	The value sequence is identical in both outputs, so their difference is exactly the operator-variation term in Theorem~\ref{thm:variation_decomposition}.
\end{proof}


\section*{Declaration of competing interest}
The authors declare that they have no known competing financial interests or personal relationships that could have appeared to influence the work reported in this paper.

\section*{Declaration of generative AI and AI-assisted technologies in the manuscript preparation process}
The generative AI tool Claude (Anthropic) was used to polish the writing of all sections, to support the retrieval and discovery of related work, to assist in checking and revising portions of the mathematical arguments in Section~\ref{sec:theory} and the proofs in \ref{app:proofs}, and to support the verification of portions of the experimental code. All AI-assisted writing was independently verified by the authors against the stated assumptions, derivations, and reported results, and all references identified with AI assistance were checked against the original publications. The authors take full responsibility for the final content of this work.



\bibliographystyle{elsarticle-num}
\bibliography{iclr2027_conference}
\end{document}